\documentclass[letterpaper, 10 pt, conference]{ieeeconf}  

\IEEEoverridecommandlockouts                              
\usepackage{amsmath}    			
\usepackage{amssymb}
\usepackage{amsfonts}
\usepackage{nccmath}
\usepackage{subcaption}
\usepackage{graphicx}  				

\usepackage[croatian]{babel}  
\usepackage[utf8]{inputenc}  	
\usepackage[T1]{fontenc}
\usepackage{ae,aecompl}     	

\usepackage{microtype}				

\usepackage{tabularx}
\usepackage{booktabs}
\usepackage{enumerate}				

\usepackage{gensymb}
\usepackage{algorithm} 
\usepackage{algpseudocode}
\usepackage{todonotes}				
\usepackage{dirtree}					
\usepackage{hyperref}					
\usepackage{color}
\usepackage{cite}
\graphicspath{{./figures/}}
\usepackage{graphicx}

\usepackage{amsthm}
\usepackage{bm}

\usepackage{subcaption}
\usepackage{graphicx}

\title{\LARGE \bf
 Towards Operating Wind Turbine Inspections using a LiDAR-equipped UAV
}
\author{Toma Sikora, Lovro Markovic, Stjepan Bogdan
	\thanks{Authors are with Faculty of Electrical and Computer Engineering,
        University of Zagreb, 10000 Zagreb, Croatia
        {\tt\small (authors) at fer.hr}}}%

\makeatletter
\newcommand{\removelatexerror}{\let\@latex@error\@gobble}

\makeatother

\begin{document}
\maketitle

\thispagestyle{empty}
\pagestyle{empty}

\begin{abstract}
In this study, a novel technique for the autonomous visual inspection of rotating wind turbine rotor blades utilizing an unmanned aerial vehicle (UAV) was developed. This approach addresses the challenges presented by the dynamic environment at hand and the requirement of maintaining a safe distance from the moving rotor blades. The application of UAV-based inspection techniques mitigates these dangers and the expenses associated with traditional wind turbine inspection methods which involve halting normal wind farm operations.

Our proposed system incorporates algorithms and sensor technologies. It relies on a light detection and ranging (LiDAR) sensor system, an inertial measurement unit, and a GPS to accurately identify the relative position of the rotating wind turbine with respect to the UAV's own position. Once this position is determined, a non-destructive visual analysis of the rotating rotor blades is performed by generating a suitable trajectory and triggering a camera fitted on a gimbal system as the blades approach. This new technique, built upon the existing research on UAV inspection of rotating wind turbines, has been empirically validated using data collected from real-world wind farm applications.
This article contributes to the ongoing trend of enhancing the safety and efficiency of infrastructure inspection. It also presents a good base for future research, with potential applications for other types of infrastructure, such as bridges or power lines.
\end{abstract}

\section{Introduction}
\label{sec:introduction}

Unmanned aerial vehicles (UAVs) have become a popular solution for performing autonomous inspection of various structures, such as windturbines. The attractive aspect of this approach is the ability to perform non destructive testing (NDT) without putting people at risk. This has led to the development of UAV-based inspection techniques in a range of tasks. Traditionally, wind turbine inspection involved either the dangerous task of personnel climbing the windturbine or in some cases the expensive task of unmounting the rotor blades for inspection.\\
In this work, a method is proposed for performing autonomous visual inspection of the rotor blades using a UAV while the turbine is rotating. Previous work on visual inspection of stationary wind turbines involves the need to halt normal wind turbine operation, making it inconvenient and expensive. Performing inspection on a rotating wind turbine, on the other hand, circumvents this problem.

Several previous studies have explored the use of UAVs, mainly quadrotors, for the inspection of stationary wind turbines. For example, vision based inspection systems of wind turbines are presented in articles such as~\cite{weibin_gu, stokkeland2015, schafer2016, ref:Car2020}.

In~\cite{weibin_gu} a vision based system for UAV wind turbine inspection is designed and the results presented both in simulation and Hardware-In-The-Loop testing. The system uses a LiDAR sensor and a camera for the vision pipeline integrating the YOLOv3 network and a customized Hough transform algorithm.\\
Furthermore, in~\cite{stokkeland2015} a machine vision module for the navigation of an UAV during the inspection of stationary wind turbines is presented. It implements the estimation of the relative position and distance between the UAV and the wind turbine, and the position of the blades. The system utilized the Hough transform for detection and the Kalman filter for tracking. Experiments show the accuracy and robustness of the solution.\\
Once again,~\cite{schafer2016} studies autonomous inspection flights at a wind turbine, focusing on the generation of an a-priori 3D map of the wind turbine, and path planning and collision avoidance algorithms. The system relies on a GPS and a 2D LiDAR sensor to collect point clouds which are used for the relative localization process.\\
Moreover, in~\cite{ref:Car2020} the team on whose work this study is based on presents a semi-autonomous wind turbine blade inspection system with a LiDAR-equipped UAV. The process performs successful wind turbine blade inspections with minimal operator involvement and results in blade images and a wind turbine 3D model. The method is tested and validated in a real life setting.\\
Lastly, in~\cite{drewry} a review of NDT techniques for wind turbines is given, giving more insight into the actual inspection techniques rather than the manner in which they are carried out (manually or automated).

When talking about the use of UAVs for the inspection of rotating wind turbines, however, the research is very limited. This is a more challenging task due to the dynamic nature of the moving turbine and the need to maintain a safe distance of the UAV from the rotating blades at all times. Nevertheless, some recent studies have shown promising results in this area.\\
For example, in~\cite{doi:10.1177/0309524X211027814} a novel method is proposed where an UAV is used to perform strain monitoring on the wind turbine rotor based on a optical digital image correlation technique. Validation of the method is performed using strain gauges proving it robust.\\
Furthermore, in~\cite{KHADKA2020106446} a digital image correlation system installed on a drone is used to obtain dynamic characteristics of the blades with high precision while they rotate, enabling on-site robust measurements without the need to stop the operation of the wind turbine.\\

Despite these advances, the following technical challenges need to be addressed to perform UAV inspection of rotating wind turbines. These are: developing robust algorithms to detect and track the blades, ensure safe and reliable operation of the UAV in the challenging and dynamic environment, and keeping track of information acquired dynamically for each blade.\\

In this article, we present the work done on the problem of rotating wind turbine inspection using a UAV. The implemented system contains the method for finding the relative position of the rotating wind turbine relative to the position of the UAV, relying heavily on the LiDAR sensor system, an inertial measurement unit, and a GPS. Moreover, once the relative position is found, we developed a method for performing non destructive visual analysis of the rotating rotor blades by planning a suitable trajectory and triggering a camera with a gimbal camera system. Having implemented the aforementioned techniques, the function of the system was experimentally tested on data collected on a real world wind farm.\\

The article is divided into sections as follows. Section \ref{sec:introduction} provides an introduction into the studied matter, including an overview of available literature focusing on similar problems. The 3D model generation method used to obtain the point cloud of the rotating wind turbine and the model matching algorithm used to find the generated model in the data stream coming from the LiDAR sensors is presented in Section \ref{sec:model_matching}. In Section \ref{sec:camera_triggering} the trajectory generation algorithm to perform the inspection the rotor blades of a rotating wind turbine and the setup allowing for autonomous camera triggering is presented. The following Section \ref{sec:experimental_results} presents the results of experiments performed using the presented approach on data collected in real world, followed by a short discussion of the results. Funally, Section \ref{sec:conclusion} contains a conclusion of the study, with a short recap of the work and possible avenues for future work.

%
\section{Model Matching} \label{sec:model_matching}

In the process of autonomous wind turbine inspection, being aware of the relative position of the wind turbine with regards to the UAVs own position is of utmost importance. Obtaining and tracking this information can be done in various ways, for example trilateration with ultrasound or ultra-wide bandwidth technologies, GPS or searching the point cloud with LiDAR. Specifically, the system in this project works on the basis of a fusion of GPS and IMU signals for global localization, and point cloud model matching for localization relative to the object being inspected, e.g. a wind turbine.\\
Model matching refers to the process of locating a specific 3D model within the point cloud captured by a sensor system, e.g. a LiDAR sensor. It involves comparing the acquired point cloud data with a reference 3D model to identify the presence and precise location of the model within the scanned area. By finding correspondences between two point clouds, one can determine the transformation (position and orientation) required to align the model with the point cloud data.\\
The rest of the section presents the techniques used to perform model matching for this setting.

\subsection{3D Model generation}
Accurate and detailed 3D models are crucial for performing inspection of rotating wind turbines. However, while for the inspection of stationary wind turbines obtaining an exact model is relatively simple, finding a rotating wind turbine in the received point clouds presents a different problem. Many approaches had been used during the development of this system, but the best results were obtained by approximating the rotor in motion with a thin cylinder/plate model. This is a geometrical representation of the trace of the wind turbine's rotor during the rotation picked up by the LiDAR sensor.\\
For the purposes of the inspection workflow, it was desirable to have the system as modular and generic as possible, so the process of model generation was scripted in a Blender Python script file. This allows parametrization with regards to the wind turbine height and rotor radius.\\
Blender is a free and open source software tool developed for 3D computer graphics. The platform is used in the creation of animated films, visual effects, artwork, 3D printable models, motion graphics, etc. by a large community. In line with its open source nature, Blender offers an open library for Python scripting, making modelling through code possible.\\
The necessary input into the script is the wind turbine pillar height, the rotor radius, and its width. The script generates a 3d model of the rotating wind turbine (a cylinder approximating the rotor) in the Blender UI. The model can easily be exported to .obj and transformed in a point cloud ready for usage.

\subsection{Matching algorithm}
\begin{figure}
    \centering
    \includegraphics[width=0.45\textwidth]{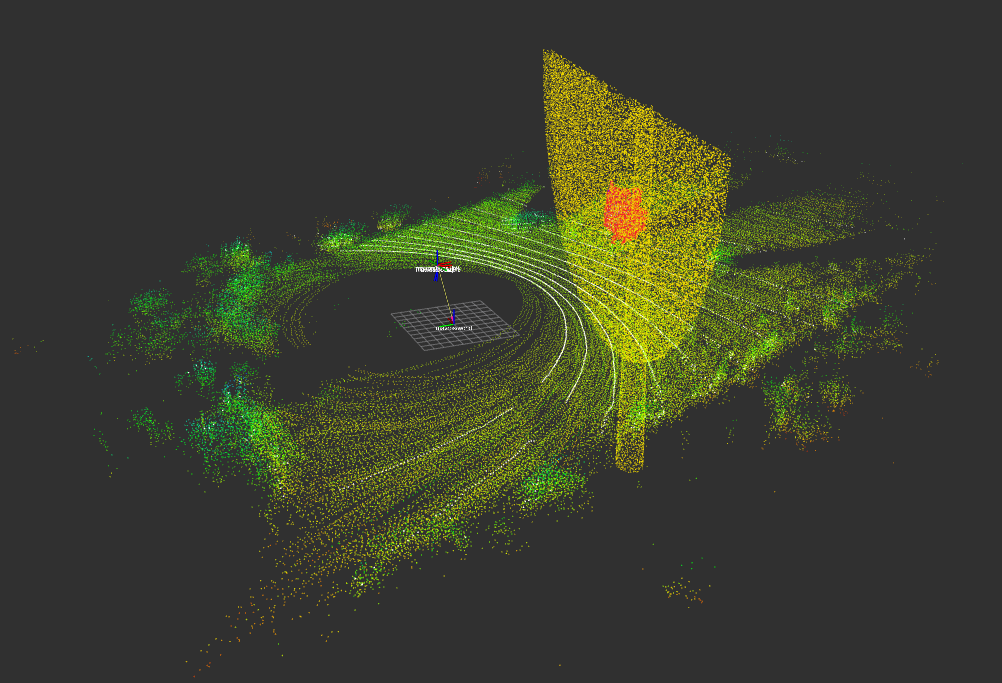}
    \caption{Matching process - Initialization}
    \label{fig:matching_1}
\end{figure}

\begin{figure}
    \centering
    \includegraphics[width=0.45\textwidth]{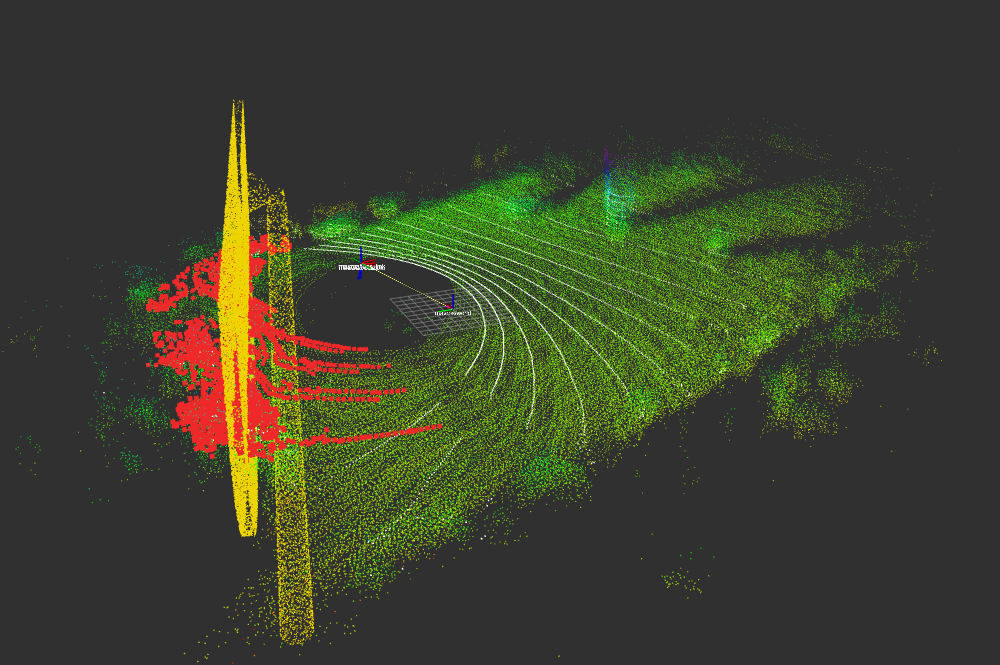}
    \caption{Matching process - First match}
    \label{fig:matching_2}
\end{figure}

\begin{figure}
    \centering
    \includegraphics[width=0.45\textwidth]{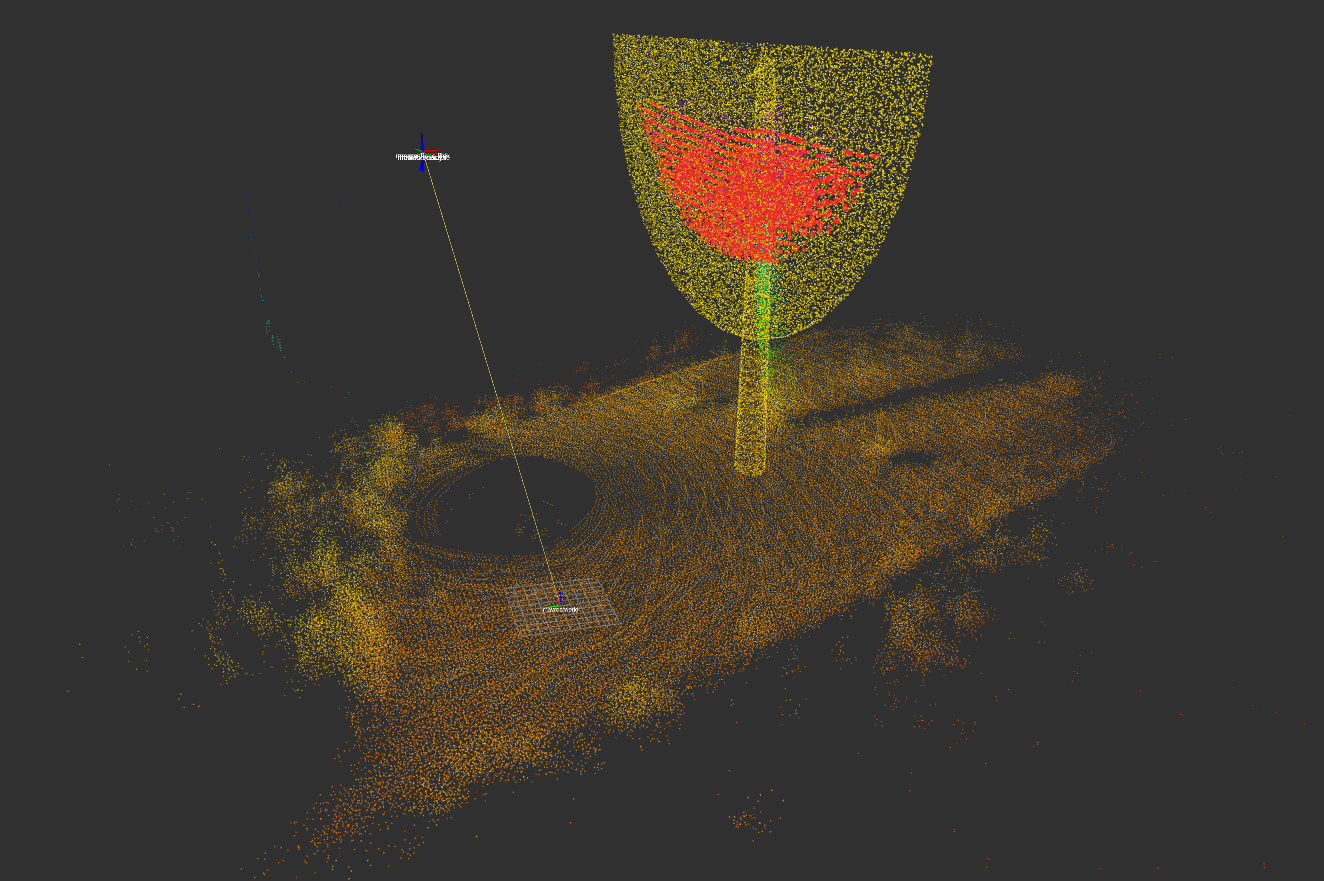}
    \caption{Matching process - Point cloud accumulation}
    \label{fig:matching_3}
\end{figure}

\begin{figure}
    \centering
    \includegraphics[width=0.45\textwidth]{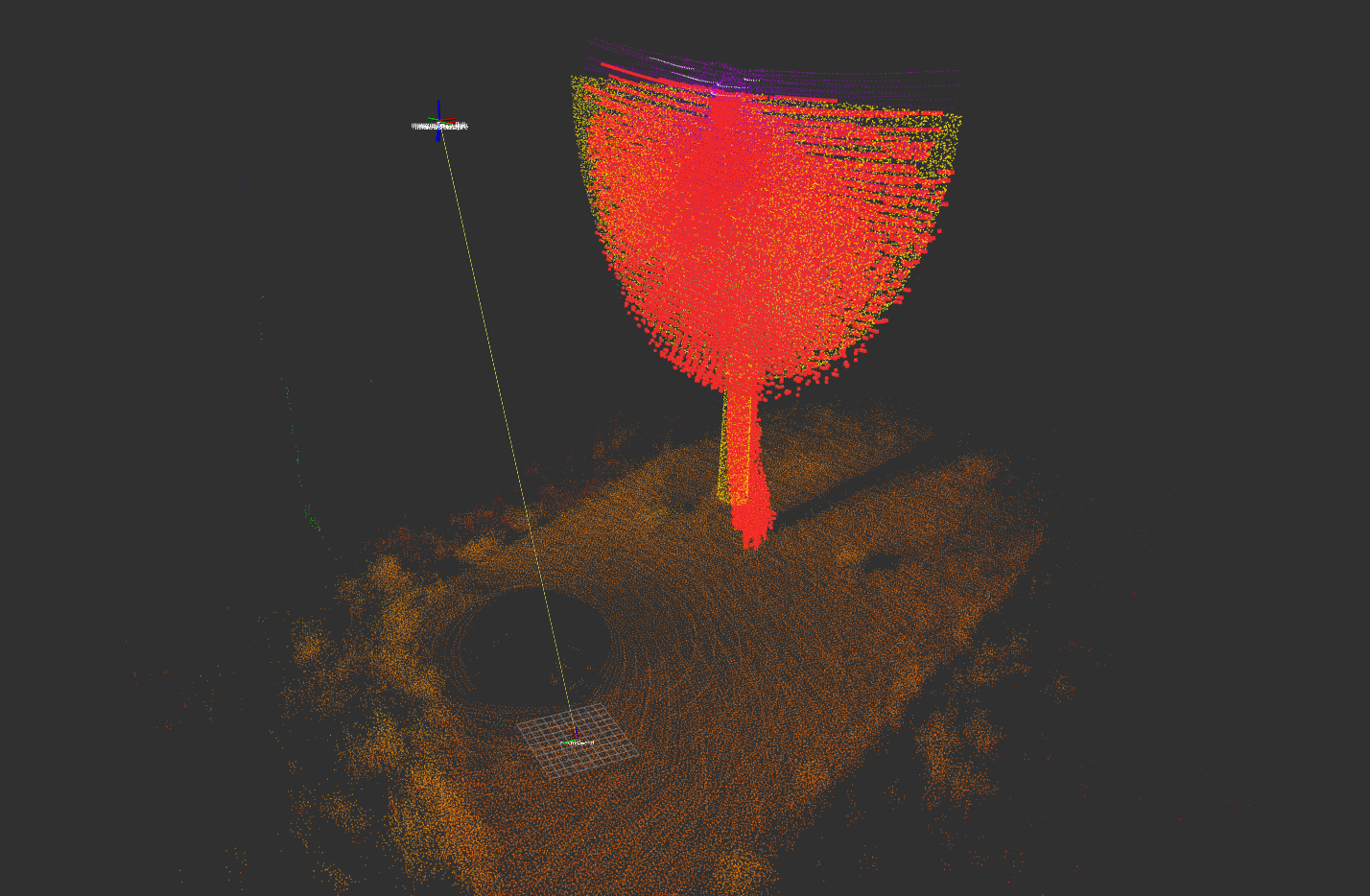}
    \caption{Matching process - Full match}
    \label{fig:matching_4}
\end{figure}

Figures~\ref{fig:matching_1},~\ref{fig:matching_2},~\ref{fig:matching_3}, and~\ref{fig:matching_4} illustrate the matching process as it appears in the RVIZ software. Figure~\ref{fig:matching_1} presents the initialized system with the wind turbine 3D model randomly positioned depicted in yellow and then fixed to the first "match" (depicted with red points) which is a group of points collected on a bush nearby in~\ref{fig:matching_2}. Figure~\ref{fig:matching_3} presents the next important situations in the matching process. Depicting the accumulated point cloud which now contains also points which are clearly from the actual wind turbine. Lastly, in figure~\ref{fig:matching_4}, once enough points coming from the wind turbine are accumulated, the matching process correctly matches the entire model (the yellow and red point clouds coincide with one another).


\begin{algorithm}
    \caption{Matching algorithm pseudocode}\label{alg:matching}
    \begin{algorithmic}[1]
        \Procedure{Rotating wind turbine matching}{}
        \State Initialize drone takeoff
        \While{unsatisfactory match}
            \State Collect point cloud
            \State Point cloud registration
            \State Point cloud voxelization
            \State Ground plane rejection
            \State Point cloud clustering
            \For{i in clusters}
                \If{found better match}
                    \State Update best match
                \EndIf
            \EndFor
        \EndWhile\\
        
        \State \Return result
        \EndProcedure
    \end{algorithmic}
\end{algorithm}

The pseudocode for the model matching process used to align and match a 3D point cloud model of a rotating wind turbine during UAV drone inspection is given with~\ref{alg:matching}. The important steps of the algorithm are the following:
\begin{itemize}
    \item \textbf{Initialization} The process is initialized by placing a UAV drone in front of the wind turbine, clearing the site, and starting the automated inspection process. The UAV drone then takes off and starts collecting point clouds as it climbs in height.
    \item \textbf{Point Cloud Registration} Every new point cloud is incorporated into the global point cloud through the process called registration.
    \item \textbf{Point Cloud Voxelization}
    \item \textbf{Ground Plane Rejection} The points in the point cloud belonging to the ground plane are removed through the ground plane rejection algorithm. The algorithm searches for a plane with most points in its neighborhood through an optimization process.
    \item \textbf{Point cloud Clustering} The remaining point cloud is then segmented into clusters based on proximity. Clustering groups together points that are spatially close to each other, representing different objects in the scene.
    \item \textbf{Match Evaluation} Iterating over the clusters, the algorithm evaluates the match with the generated 3d model of the rotating wind turbine. The evaluation is done with the Iterative Closest Point (ICP) algorithm, keeping the pitch and roll values constant. This follows the assumption that the UAV drone stays upright during inspection.
\end{itemize}

Once the UAV drone reaches the desired height at which the inspection process can begin, the best match the algorithm found should very closely represent the actual position and orientation of the wind turbine relative to the drone.

\section{Camera Triggering} \label{sec:camera_triggering}

\subsection{Trajectory generation}
Having finished with the most critical part of the inspection, locating the wind turbine in the local reference frame, it is necessary to plan the trajectory with regards to its position. The generated trajectories allow the UAV drone to capture images of points of interest on the rotating blades.

The generation process involves a Python script which produces trajectories and stores them in a .csv file format. The script is designed to generate a trajectory relative in position to the wind turbine, taking into account the specific parameters of the wind turbine itself. These are its base height and rotor radius. The Python script's flexibility allows for easy adjustment and adaptation through parametrization, ensuring that the drone can navigate the inspection process effectively and efficiently on virtually any model of the turbine.

Through the generated trajectory, the inspection encompasses the front and back sides of the rotor blade seen from three different angles. To capture comprehensive visual information, the drone's trajectory should go as follows:
\begin{enumerate}
    \item Approach the rotor center head-on at a safe distance.
    \item Move sideways to the tip of the blades.
    \item Make a 180\degree turn around the tip to reach the back side of the rotor.
    \item Move sideways to the rotor center behind the turbine.
    \item Repeat with a height offset above and below the traversed line, pointing the gimbal at 45 degrees.
\end{enumerate}

\begin{figure}
    \centering
    \includegraphics[width=0.45\textwidth]{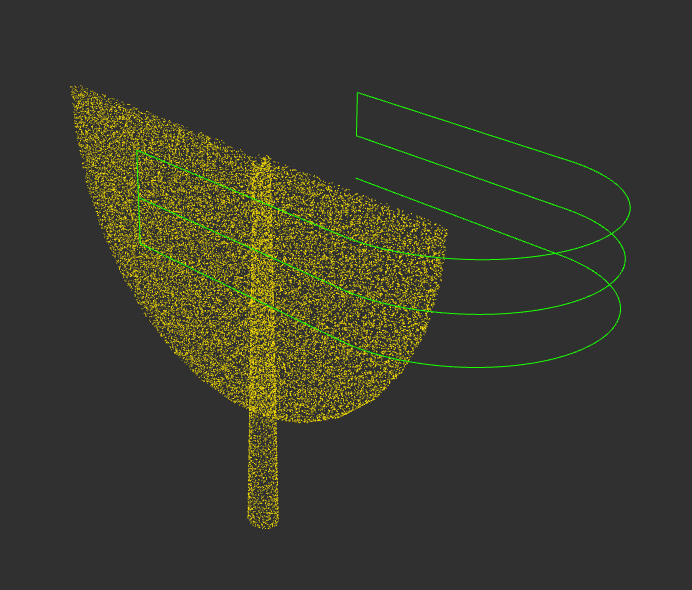}
    \caption{Generated rotating wind turbine model with the generated trajectory}
    \label{fig:trajecotry_model}
\end{figure}

Figure~\ref{fig:trajecotry_model} depicts the generated point cloud of the 3d model representing the wind turbine and the trajectory generated with the adequate parameters.

This trajectory generation approach enhances the efficiency and accuracy of wind turbine inspections by streamlining the data collection process. The collected images can subsequently be analyzed and processed for analysis, such as detecting potential defects or monitoring the condition of the wind turbine.

\subsection{Camera Triggering}

To capture images of the rotating blades during autonomous wind turbine inspection using a UAV drone, a precise camera triggering process has to be created.

\begin{figure}
    \centering
    \includegraphics[width=0.45\textwidth]{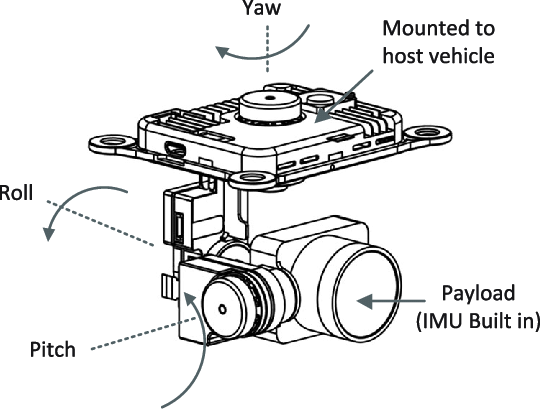}
    \caption{An example of a gimbal system for a commercial UAV drone.  Source image from \cite{Fu-Gimbal}.}
    \label{fig:gimbal}
\end{figure}

The image capture process has to be synchronized with the passing of the blades. To ensure that images are taken at the right time, the camera triggering mechanism contains a camera and a laser range finder fitted onto a gimbal, an example of which is given in~\ref{fig:gimbal}. This gimbal is used to point the two in the desired direction and an Arduino micro-controller is used to connect the laser range finder to the camera trigger.

As the blade approaches while the UAV is waiting in position, the laser range finder detects the blade's presence in front of it, signaling the Arduino micro-controller to activate the camera trigger. Upon receiving the signal from the laser range finder, the camera trigger initiates the image capture process, capturing a picture of the rotor blade from the desired angle. This synchronized approach guarantees that the images correspond to the targeted points of interest, providing valuable data for further analysis.

\section{Experimental Results} \label{sec:experimental_results}

This section presents the results of the experiments conducted with the implemented autonomous inspection process for a rotating wind turbine. Firstly, the details on the experiment setting will be introduced, followed by the results obtained through the experiments. Lastly, the section will end with a small discussion on the matter.

The work done in this study relies heavily on the code bases for autonomous UAV operation and point cloud registration and inspection developed by the LARICS \footnote{Laboratory for Robotics and Intelligent Control Systems - \hyperlink{https://larics.fer.hr/}{larics.fer.hr}} laboratory from the University of Zagreb, similar to the work done in \cite{ref:Car2020}. The autonomous UAV operation code base takes care of the low level control providing an interface for higher level navigation and inspection scheduling. On the other hand, the point cloud registration and inspection code base provides high level handling of point cloud data coming from the LiDAR sensors, including operations such as registration, ground plane filtering, clustering, and matching.

The data used to perform experiments was collected by a custom UAV equipped with an array of sensors. The system runs on an Intel NUC with ROS as the communication base. Moreover, the flight control is done through the Pixhawk and Ardupilot firmware. The positioning is done through a fusion of GPS and IMU signals, and based on the run, the UAV is fitted with either a Velodyne or an Ouster LiDAR sensor.

\begin{table}[H]
    \centering
    \begin{tabular}{||c|c||}
    \hline
     \multicolumn{2}{||c||}{Pometeno Brdo wind farm} \\
     \hline
    \hline
        Location & near Split, Croatia \\
    \hline
        Base height & 45 m \\
    \hline
        Rotor radius & 30 m \\
    \hline
        Turbine number & 17 \\
    \hline
        Electricity generation & 84,7 MW \\
    \hline
    \end{tabular}
    \caption{Inspected wind turbine information}
    \label{tab:turbine_info}
\end{table}

The data collection process was done on a wind turbine farm near Split, Croatia with details given in~\ref{tab:turbine_info}. It involved bringing the UAV up in front of a wind turbine and hovering in front of it for an amount of time, all the while collecting information about the environment with the LiDAR. The data is stored ROS .bag files for playback and further experiments in simulation. This data was used to validate the process of localization of the rotating turbine with the algorithm implemented in this study.

\begin{table}[H]
    \centering
    \begin{tabularx}{0.48\textwidth}{||X|X|X|X||}
    \hline
     Run no. & LiDAR & Match results & Yaw (rad) \\
     \hline
    \hline
        1 & Ouster & 0.262016 & 1.73169\\
    \hline
        2 & Ouster & 0.251234 & 1.98284\\
    \hline
        3 & Ouster & 0.277545 & 2.187747\\
    \hline
    \end{tabularx}
    \caption{Experimental results}
    \label{tab:results}
\end{table}

The results of the experiments are given in table~\ref{tab:results}. The experiments are performed with the LiDAR sensor Ouster. The principal dimensionless metric for the matching process is the match score between one of the clusters in the registered point cloud and the 3d model presenting the wind turbine. The value of the match score decreases with higher match quality. Furthermore, the yaw value representing the yaw angle of the matched wind turbine in the scene. Generally, for clusters not resembling the wind turbine the match score is in the 100-200 range, for clusters similar to a part of the turbine 1-100, and for a good match a score <1 is expected. \\
Let us discuss the matching results for the Ouster LiDAR (the first three rows in the table). These represent experiments done on January 25th 2022 and as can be seen from the match results, all of them successfully matched the 3d model onto the registered point cloud. The score of all three experiments is in the 0.25-0.28 range, resulting in an almost perfect fit. This can be further confirmed visually with matches similar to~\ref{fig:matching_4}. Moreover, from yaw values being different for every run, it is clear that the system was initialized at a different points with regards to the wind turbine.\\ 

\section{Conclusion}
\label{sec:conclusion}

This study presents the successful development and implementation of an autonomous system for non-destructive inspection of rotating wind turbines, utilizing a UAV drone. Firstly, the article outlines the topic of research and provides an extensive examination of relevant literature, laying a solid theoretical foundation. Secondly, a presentation of the implemented model matching algorithm and the 3D model generation process is given, reflecting a progressive design approach to cater to the requirements of the inspection setting. The model matching algorithm works on a data-stream coming in from a LiDAR sensor mounted on board, resulting in a approximation of the relative wind turbine position with regards to the UAV. The 3D model generation process allows for a modular and fast re-purposing based on the turbine at hand. We then detailed the technical aspects of the parametrized trajectory generation algorithm and the camera triggering system to obtain high resolution images of the critical rotor blade parts. Finally, the proposed system underwent a series of tests using real-world data, validating its practicality and effectiveness.

The work can be further developed a number of future directions, such as the application of this methodology to other infrastructure, e.g. bridges and towers. This area of possible expansion of applicability underlines the versatility of the system. Additionally, a higher level of parametrization in the inspection process can be explored as another potential advancement, which could refine the system further and increase its efficiency. This research, therefore, contributes to the ongoing development of autonomous non-destructive inspections using UAVs.

\section*{ACKNOWLEDGMENT}
This research was supported by European Commission Horizon 2020 Programme through project under G. A. number 820434, named ENergy aware BIM Cloud Platform in a COst-effective Building REnovation Context - ENCORE \cite{ENCOREweb}. Furthermore, this research was a part of the scientific project Autonomous System for Assessment and Prediction of infrastructure integrity (ASAP) \cite{ASAPweb} financed by the European Union through theEuropean Regional Development Fund-The Competitiveness and Cohesion Operational Programme (KK.01.1.1.04.0041). The work of doctoral student Marko Car has been supported in part by the “Young researchers’ career development project--training of doctoral students” of the Croatian Science Foundation funded by the European Union from the European Social Fund.






\bibliographystyle{ieeetr}
\bibliography{bibliography/Mendeley}

\end{document}